\documentclass{article}

\usepackage{microtype}
\usepackage{graphicx}
\usepackage{subcaption}
\usepackage{booktabs} 

\usepackage{hyperref}
\usepackage{multirow}
\usepackage{makecell}

\usepackage[preprint]{icml2026}

\usepackage{amsmath}
\usepackage{amssymb}
\usepackage{mathtools}
\usepackage{amsthm}

\usepackage[capitalize,noabbrev]{cleveref}

\theoremstyle{plain}

\theoremstyle{definition}

\theoremstyle{remark}

\usepackage[textsize=tiny]{todonotes}

\icmltitlerunning{Beyond Open Vocabulary: Multimodal Prompting for Object Detection in Remote Sensing Images}

\begin{document}

\twocolumn[
  \icmltitle{Beyond Open Vocabulary: Multimodal Prompting for \\ Object Detection in Remote Sensing Images
}



  \icmlsetsymbol{equal}{*}

 \begin{icmlauthorlist}
    \icmlauthor{Shuai Yang}{sch}
    \icmlauthor{Ziyue Huang}{sch}
    \icmlauthor{Jiaxin Chen}{sch}
    \icmlauthor{Qingjie Liu}{sch}
    \icmlauthor{Yunhong Wang}{sch}
  \end{icmlauthorlist}

  \icmlaffiliation{sch}{School of Computer Science and Engineering, Beihang University, Beijing, China}

\icmlcorrespondingauthor{Qingjie Liu}{qingjie.liu@buaa.edu.cn}

  \icmlkeywords{Machine Learning, ICML}

  \vskip 0.3in
]



\printAffiliationsAndNotice{}  

\begin{abstract}
Open-vocabulary object detection in remote sensing commonly relies on text-only prompting to specify target categories, implicitly assuming that inference-time category queries can be reliably grounded through pretraining-induced text--visual alignment.
In practice, this assumption often breaks down in remote sensing scenarios due to task- and application-specific category semantics, resulting in unstable category specification under open-vocabulary settings.
To address this limitation, we propose \textbf{RS-MPOD}, a multimodal open-vocabulary detection framework that reformulates category specification beyond text-only prompting by incorporating instance-grounded visual prompts, textual prompts, and their multimodal integration.
RS-MPOD introduces a visual prompt encoder to extract appearance-based category cues from exemplar instances, enabling text-free category specification, and a multimodal fusion module to integrate visual and textual information when both modalities are available.
Extensive experiments on standard, cross-dataset, and fine-grained remote sensing benchmarks show that visual prompting yields more reliable category specification under semantic ambiguity and distribution shifts, while multimodal prompting provides a flexible alternative that remains competitive when textual semantics are well aligned.

\end{abstract}

\begin{figure}[t]
    \centering
    \includegraphics[width=0.98\linewidth]{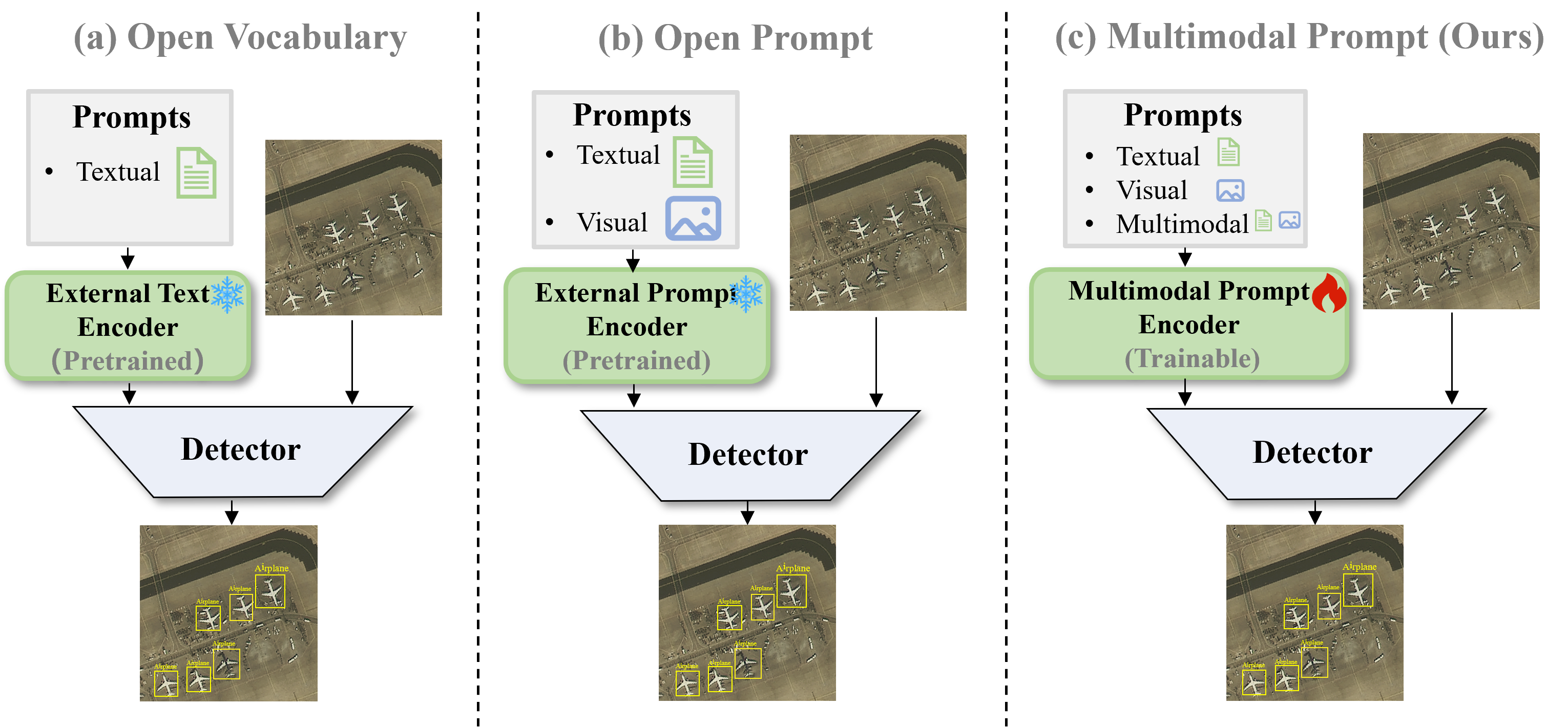}
    \caption{\textbf{Prompting paradigms for object detection.}
    (a) Open-vocabulary prompting only supports textual prompts.
    (b) Open prompt extends prompts to textual and visual inputs via an external pre-trained prompt encoder.
    (c) Our multimodal prompting jointly supports textual, visual, and multimodal prompts with trainable prompt encoders.
    }
    \label{fig:introduction}
\end{figure}

\section{Introduction}
Object detection is a fundamental task in remote sensing image analysis, where models are required to localize objects and assign category labels under large variations in spatial resolution, sensing conditions, and background complexity~\cite{cheng2016survey, li2020survey}.
With the rapid development of remote sensing imaging technologies~\cite{xia2018dota, li2020dior} and deep learning architectures~\cite{he2016resnet, ren2016fasterrcnn, vaswani2017transformer}, modern detectors have achieved strong performance across a wide range of challenging remote sensing scenarios~\cite{xie2021orientedrcnn, yang2021r3det, lyu2022rtmdet, zeng2024arsdetr}.
Despite these advances, most existing detectors remain constrained to fixed and closed category sets, which fundamentally limits their ability to recognize objects beyond the predefined training taxonomy.

To relax this limitation, the computer vision community has developed open-vocabulary object detection (OVOD)~\cite{zareian2021openvocabulary}, which formulates category prediction as a similarity-based matching problem between region-level visual representations and category embeddings.
In practice, most OVOD methods derive category embeddings from category names or textual descriptions, and rely on large-scale image--text pretraining~\cite{li2022glip, liu2024gdino, cheng2024yoloworld} or supervision from vision--language models~\cite{gu2021vild, zhao2022vlplm} to establish cross-modal correspondence, enabling category specification at inference time through textual prompts.

Recent studies have begun to extend open-vocabulary object detection to the remote sensing domain~\cite{li2024castdet, wei2024ovadetr}, aiming to relax the closed-set constraint imposed by fixed category taxonomies.
However, most existing approaches continue to rely almost exclusively on text-based category specification, implicitly assuming that inference-time category names can be reliably grounded to stable and discriminative visual concepts within the text--visual embedding space learned during vision--language pretraining (Fig.~\ref{fig:introduction}(a)).
In remote sensing scenarios, this assumption is often violated.
Unlike natural image benchmarks, category vocabularies and naming conventions in remote sensing are highly task- and application-dependent, and frequently diverge from those reflected in large-scale pretraining corpora.
As a result, inference-time textual prompts may correspond to imprecise or biased visual concepts, making pretraining-induced text--visual alignment insufficient for reliable category specification.

This issue is further amplified under open-vocabulary settings, where category specification relies almost entirely on textual prompts without task-specific supervision, often leading to unstable or inconsistent detection behavior across datasets and category granularities.
To alleviate the limitations of text-only prompting, recent work has explored incorporating visual exemplars as an alternative form of category specification~\cite{huang2025openrsd}, by encoding exemplar images as visual prompts using external pretrained models (Fig.~\ref{fig:introduction}(b)).
However, such approaches typically treat textual and visual prompts as independent, unimodal inputs, and require additional alignment steps to project externally encoded visual prompts into the detector feature space.
Moreover, they do not explicitly account for the potentially complementary information across different prompt modalities.

To address these limitations, we propose RS-MPOD, a multimodal prompting framework for open-vocabulary object detection that rethinks category specification beyond text-only prompting.
RS-MPOD enables category specification through textual prompts, instance-grounded visual prompts, or their multimodal combination within a unified prompt-based detection framework (Fig.~\ref{fig:introduction}(c)).
RS-MPOD is built upon GroundingDINO~\cite{liu2024gdino}, which provides a generic prompt-based detection architecture.
Our design centers on category specification in prompt-based open-vocabulary detection, where visual features are encoded independently of category information and all category cues are introduced solely through prompt embeddings during decoding.
To this end, we introduce a visual prompt encoder based on deformable attention~\cite{zhu2020deformableattention} to extract instance-grounded appearance representations from annotated exemplar objects.
These instance-level visual prompts provide a text-free mechanism for specifying categories based on visual similarity.
At inference time, multiple visual prompts can be flexibly aggregated to form category-level representations, enabling open-vocabulary detection conditioned directly on visual exemplars.
When both textual and visual cues are available, we further introduce a multimodal prompt fusion module that integrates prompts from different modalities into unified category representations, allowing the detector to exploit complementary semantic and appearance information.

Our contributions can be summarized as follows.
(1) We identify the limitations of text-only category specification in remote sensing open-vocabulary object detection, and extend category specification to support visual exemplars and multimodal cues beyond category names.
(2) We introduce a visual prompt encoder that extracts instance-grounded visual prompts from annotated exemplar objects to enable appearance-based category specification, together with a multimodal prompt fusion module that combines textual semantics with visual appearance cues to form unified category representations.
(3) Extensive experiments on standard, cross-dataset, and fine-grained remote sensing benchmarks show that visual and multimodal category specification improves robustness under semantic ambiguity and distribution shifts, while remaining competitive when textual semantics are well aligned.

\begin{figure*}[t]
  \centering
  \includegraphics[width=0.98\linewidth]{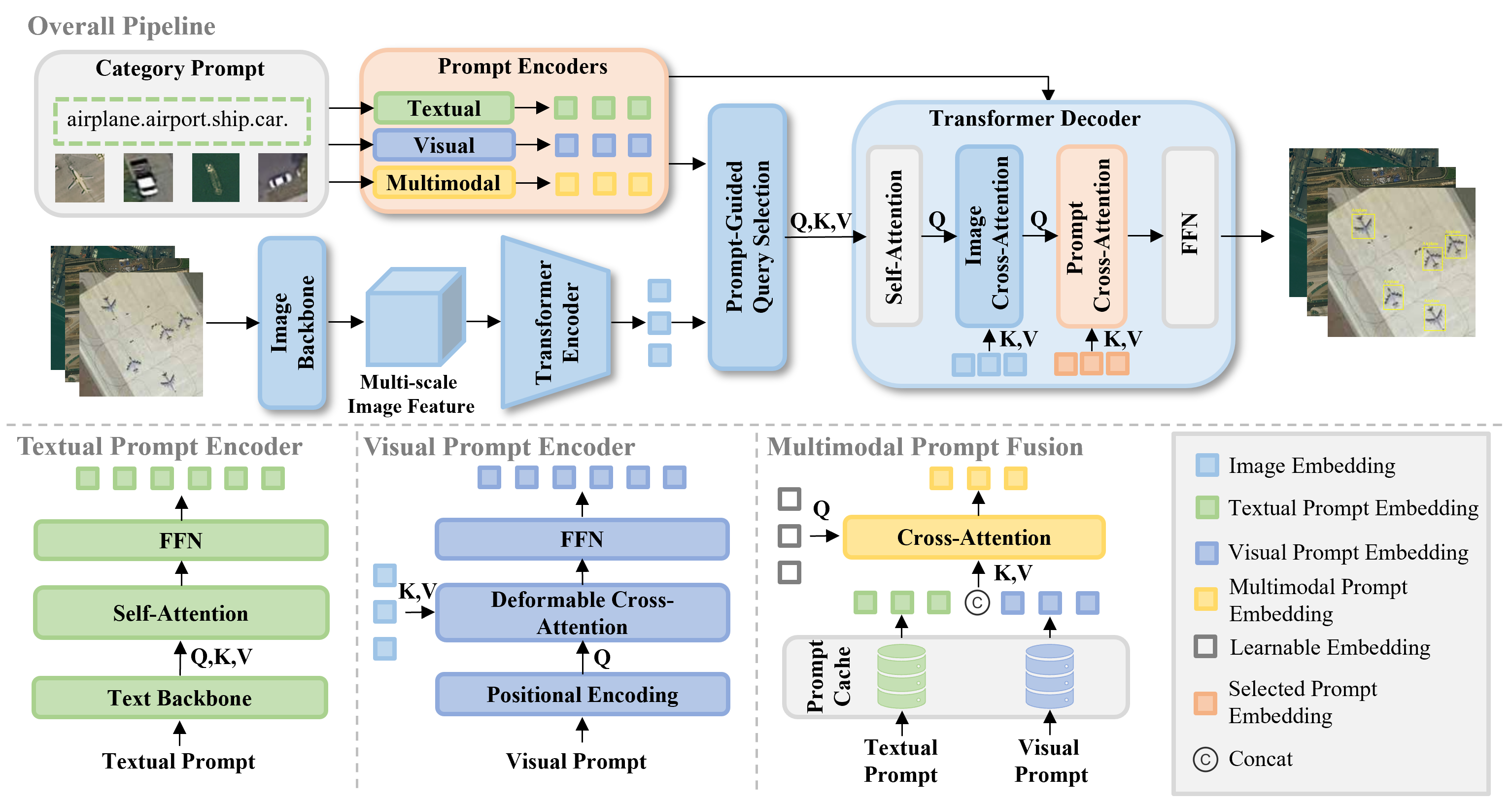}
  \caption{\textbf{Overall framework of RS-MPOD.}
The detector is built upon GroundingDINO and supports textual prompting, visual prompting, and multimodal prompting within a unified detection pipeline.
Prompt embeddings produced by different prompt encoders are used to condition query selection and cross-attention in the transformer decoder for category specification.
The lower panels illustrate the designs of the textual prompt encoder, visual prompt encoder, and the multimodal prompt fusion module.}
  \label{fig:overview}
\end{figure*}

\section{Related Work}
\subsection{OVOD in Natural Images}
Open-vocabulary object detection (OVOD) casts category recognition as a similarity-based matching problem between region-level visual features and textual embeddings, enabling detectors to recognize categories beyond a fixed training taxonomy.
Existing OVOD approaches in natural images can be broadly categorized into two paradigms according to how such cross-modal alignment is established.

The first paradigm explicitly incorporates cross-modal fusion mechanisms into the detector architecture and learns vision--language alignment through end-to-end training on large-scale image--text pairs \cite{li2022glip, yao2022detclip, minderer2022owlvit, liu2024gdino, cheng2024yoloworld}.
These methods directly couple visual and textual streams during training, enabling strong cross-modal alignment, at the cost of increased reliance on large, well-curated image--text datasets and extensive training supervision.

The second paradigm builds upon pretrained vision--language models (VLMs) \cite{radford2021clip, jia2021align, li2022blip}, which provide a shared semantic embedding space learned from large, web-scale image--text corpora.
Within this paradigm, some approaches distill alignment knowledge from VLMs into detection models \cite{gu2021vild, ma2022hierkd, wang2023oadp, wu2023clipself, fu2025hdovd, cao2025opendet}, while others exploit VLMs to generate pseudo-labels for unlabeled images \cite{zhao2022vlplm, gao2022pbovd, wang2024marvelovd, zhao2024sasdet}, thereby transferring semantic structure into detection models through indirect supervision.
While these paradigms have proven effective for open-vocabulary detection in natural images, they typically rely on the implicit assumption that the vision--language alignment learned during pretraining or end-to-end training can be directly reused to support category specification at inference time.

\subsection{OVOD in Remote Sensing Images}
Several recent studies have explored extending open-vocabulary object detection to remote sensing imagery.
OVA-DETR \cite{wei2024ovadetr} enhances vision--language alignment by introducing cross-modal fusion between visual and textual features within the detector.
CastDet \cite{li2024castdet} leverages Remote-CLIP \cite{liu2024remoteclip} as a teacher model to generate pseudo-labels for unlabeled images, thereby expanding the effective label space for open-vocabulary detection in remote sensing.
To further increase data diversity, LAE-DINO \cite{pan2025laedino} constructs a large-scale remote sensing image--text corpus for pretraining, exposing detection models to a broader range of category descriptions.
Despite these advances, most existing remote sensing OVOD methods primarily follow text-centric prompting and alignment strategies developed for natural images.
Accordingly, category specification at inference time relies mainly on textual prompts together with vision--language alignment learned during pretraining.
When applied to remote sensing datasets, variations in category definitions and usage requirements across tasks can make such text-based category specification less reliable.

Beyond textual prompting, only a limited number of studies have explored the use of visual prompts in remote sensing.
For instance, Huang et al.~\cite{huang2025openrsd} employ visual prompts extracted offline using external pretrained models.
However, visual prompts in these approaches are treated independently from textual prompts and require additional alignment to the detector feature space, without explicitly modeling interactions between different prompt modalities.

\section{Method}

\subsection{Overview}
The overall architecture of the proposed framework is illustrated in Fig.~\ref{fig:overview}.
Our detector is built upon GroundingDINO~\cite{liu2024gdino}, where visual features are extracted by an image backbone and transformer encoder, and category information is provided through prompt embeddings.
Depending on the prompting configuration, prompt embeddings are generated by different prompt encoders and guide the detection process by conditioning query selection and cross-attention decoding, resulting in category and bounding box predictions.

The framework supports three prompting configurations within a unified detection pipeline.
Textual prompting encodes category names into textual prompt embeddings.
Visual prompting specifies categories using exemplar instances from annotated training data, which are processed by a visual prompt encoder to obtain instance-level appearance representations reused at inference time.
Multimodal prompting further integrates textual and visual prompt embeddings through a multimodal prompt fusion module.

\subsection{Prompt-Based Detection Framework}

We build our detector upon GroundingDINO~\cite{liu2024gdino} and consider open-vocabulary object detection under a prompt-conditioned detection framework.
Given an input image $I$ and category prompts $\mathcal{P}=\{P_k\}_{k=1}^{K}$, visual features are extracted by an image backbone as $F=\{F^{(l)}\}_{l=1}^{L}=B_v(I)$ and further processed by a transformer encoder to obtain encoded features $\tilde{F}=E_v(F)$.
Conditioned on the category prompts, a set of queries $\{q_i\}$ is selected from the encoded visual features, which we denote abstractly as $\{q_i\}=\Phi(\tilde{F},\mathcal{P})$, where $\Phi(\cdot)$ denotes a prompt-conditioned query initialization operation.
These queries are then refined by a transformer decoder $D(\cdot)$, producing category-aware query representations $\hat{q}_i=D(q_i,\tilde{F})$ that are used for object classification and localization.

Each category $k$ is specified by a prompt embedding $P_k$, and a decoded query $\hat{q}_i$ is classified by comparing it against the category prompts.
Bounding boxes are predicted from the decoded queries by an MLP head, i.e., $\hat{b}_i=h_{\text{box}}(\hat{q}_i)$.

Following a DETR-style formulation with bipartite matching, predictions are matched to ground-truth objects, and we denote by $N$ the number of matched pairs.
The classification loss is defined as:
\begin{equation}
\mathcal{L}_{\text{cls}}
= -\frac{1}{N}\sum_{i=1}^{N}
\log
\frac{
\exp\!\big(\cos(\hat{q}_i,P_{y_i})/\tau\big)
}{
\sum_{k=1}^{K}\exp\!\big(\cos(\hat{q}_i,P_k)/\tau\big)
},
\end{equation}
where $y_i$ is the ground-truth category label of the matched object and $\tau$ is a temperature parameter.
Bounding boxes are supervised using $\mathcal{L}_{1}$ and GIoU losses:
\begin{equation}
\mathcal{L}_{1}=\frac{1}{N}\sum_{i=1}^{N}\big\lVert \hat{b}_i-b_i\big\rVert_{1},
\end{equation}
\begin{equation}
\mathcal{L}_{\text{giou}}=\frac{1}{N}\sum_{i=1}^{N}\big(1-\mathrm{GIoU}(\hat{b}_i,b_i)\big),
\end{equation}
where $b_i$ denotes the ground-truth bounding box corresponding to the $i$-th matched prediction.
The overall training objective is:
\begin{equation}
\mathcal{L}
= \lambda_{\text{cls}}\,\mathcal{L}_{\text{cls}}
+ \lambda_{1}\,\mathcal{L}_{1}
+ \lambda_{\text{giou}}\,\mathcal{L}_{\text{giou}}.
\end{equation}
where $\lambda_{\text{cls}}$, $\lambda_{1}$, and $\lambda_{\text{giou}}$ are scalar hyperparameters that balance the classification and box regression losses.

Within the prompt-based detection framework described above, we first describe the textual prompting setting, which follows the original GroundingDINO implementation.

Category specifications are provided as a set of textual prompts
$T=\{t_k\}_{k=1}^{K}$, where each $t_k$ denotes a category name.
Each textual prompt is tokenized into $n_k$ tokens and processed by a textual prompt encoder, consisting of a text backbone followed by self-attention and feed-forward layers.
The encoder outputs token-level textual features
$G_k=\{g_{k,j}\}_{j=1}^{n_k}$, which serve as the textual prompt embeddings for category $k$.
For consistency with the prompt-conditioned detection framework, we denote the category prompt for textual prompting as $P_k=G_k$.

Compared to the original GroundingDINO, we remove the feature enhancement module in transformer encoder that injects textual information into visual encoding, so that visual features are encoded independently of prompt modalities and category information is introduced solely through the prompt set $\mathcal{P}$ without modifying the detection pipeline.

\begin{figure*}[t]
  \centering
  \includegraphics[width=0.98\linewidth]{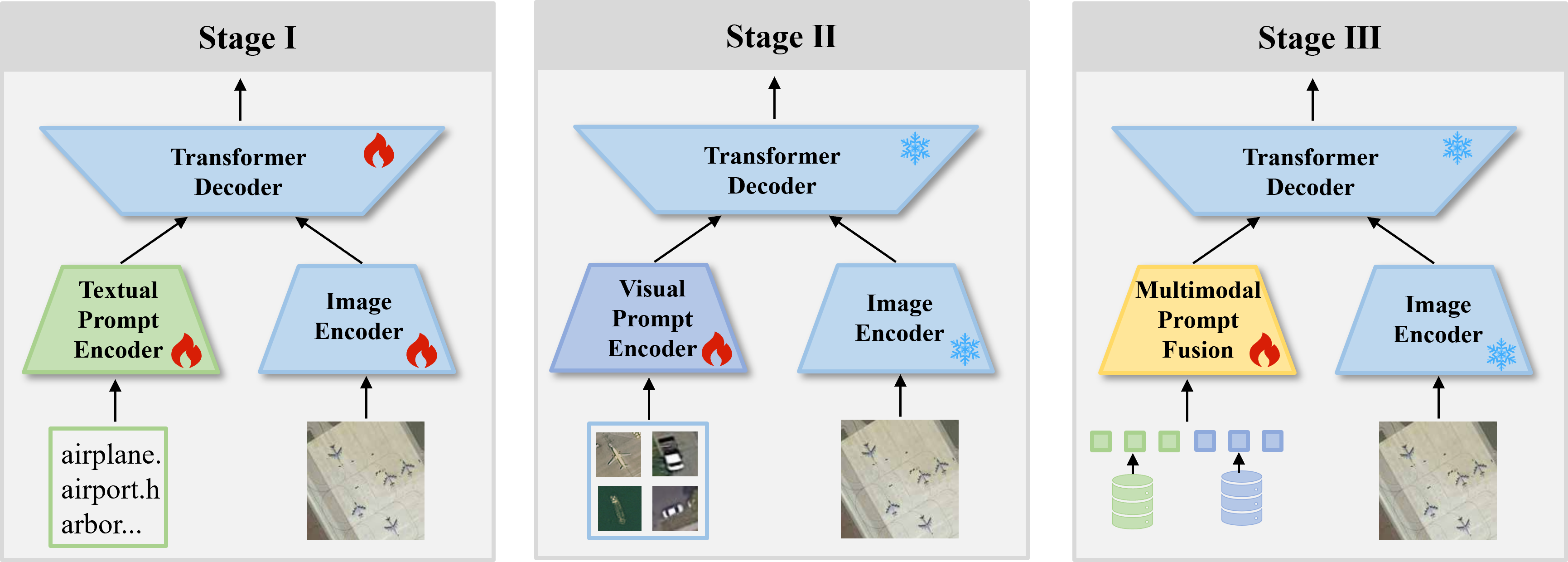}
  \caption{\textbf{Stage-wise training strategy of RS-MPOD.}
The detector is first trained with textual prompts, followed by visual prompt encoder training and multimodal prompt fusion, with earlier components frozen in later stages.
The image encoder consists of an image backbone and a transformer encoder.
}
  \label{fig:stage-wise-training}
\end{figure*}

\subsection{Visual Prompt Encoder}

Inspired by TRex2~\cite{jiang2024trex2}, we introduce a visual prompt encoder that constructs visual prompts from annotated object instances.
These instance-grounded visual prompts serve as category prompts in the prompt-based detection framework, allowing object detection to be conditioned on visual appearance.

Given the encoded multi-scale visual features $\{\tilde{F}^{(l)}\}_{l=1}^{L}$ produced by the image backbone and transformer encoder, we construct visual prompts corresponding to categories present in the image.
For an image containing a set of categories $\mathcal{C}_I \subseteq \{1,\dots,K\}$, we randomly sample one ground-truth bounding box $b_k$ for each $k \in \mathcal{C}_I$.
Each bounding box serves as a spatial prior for extracting visual information associated with category $k$.

The bounding box $b_k$ is encoded by a positional encoding function and projected into the prompt space via an MLP:
\begin{equation}
e_k=\mathrm{MLP}(\mathrm{PE}(b_k)).
\end{equation}
We introduce a learnable content vector $c_k$ and concatenate it with the positional embedding to form a prompt query:
\begin{equation}
r_k=[e_k; c_k].
\end{equation}
Each prompt query $r_k$ aggregates visual information from the multi-scale features through deformable attention:
\begin{equation}
z_k=\mathrm{DeformAttn}(r_k,\{\tilde{F}^{(l)}\}_{l=1}^{L}),
\end{equation}
and is processed by a feed-forward network to obtain the visual prompt embedding:
\begin{equation}
v_k=\mathrm{FFN}(z_k).
\end{equation}
The resulting embedding $v_k$ serves as the visual prompt for category $k$, and we denote the corresponding category prompt as $P_k = v_k$.
After training, the visual prompt encoder is applied to the training set to extract instance-level visual prompts for all annotated objects.
For each category, the extracted visual prompts are collected into a category-specific cache.
At inference time, a visual category prompt is obtained by sampling an arbitrary number of instance-level prompts from the corresponding cache and aggregating them by averaging, and the resulting representation is used as the visual category prompt $P_k$ in the detection framework.

\subsection{Multimodal Prompt Fusion Module}

We introduce a multimodal prompt fusion module to combine textual and visual prompts within the prompt-based detection framework.
In textual prompting, category prompts are represented as token-level feature sequences, whereas in visual prompting each category is represented by a single embedding.
To integrate these heterogeneous representations, we adopt a lightweight fusion mechanism based on a single-layer multi-head cross-attention module.

For each category $k$, the textual prompt encoder produces token-level textual features
$G_k=\{g_{k,j}\}_{j=1}^{n_k}$, while the visual prompt encoder produces a visual prompt embedding $v_k$.
We introduce a learnable query vector $u_k$ to aggregate information from both modalities.
The textual tokens and the visual prompt are concatenated along the token dimension to form the key/value sequence:
\begin{equation}
S_k=[\,g_{k,1},\ldots,g_{k,n_k};v_k\,].
\end{equation}
Cross-attention is applied with $u_k$ as the query and $S_k$ as both key and value:
\begin{equation}
\hat{u}_k=\mathrm{CrossAttn}(u_k,S_k,S_k).
\end{equation}
The fused embedding $\hat{u}_k$ serves as the category prompt in the multimodal prompting setting, and we denote $P_k=\hat{u}_k$.

\subsection{Stage-wise Training Strategy}
\label{sec:stagewise-training}

We adopt a three-stage training strategy (Fig.~\ref{fig:stage-wise-training}) to decouple detector learning from prompt optimization: the detector learns general-purpose visual representations, while prompt-related modules encode category-specific cues from different modalities.
This design avoids interference between prompt learning and stable detection feature formation.

\noindent\textbf{Stage I: Detector training with textual prompts.}
We first train the detector with textual prompts to establish a stable detection backbone.
All detector components and the textual prompt encoder are optimized in this stage.
The resulting model provides shared visual representations for subsequent prompt learning.

\noindent\textbf{Stage II: Visual prompt encoder.}
In the second stage, the detector trained in Stage~I is frozen, and only the visual prompt encoder is optimized.
After training, the visual prompt encoder is applied to the training set to construct category-specific visual prompt caches, which are used for visual and multimodal prompting in later stages.

\noindent\textbf{Stage III: Multimodal prompt fusion module.}
In the final stage, the detector and prompt encoders remain fixed.
The fusion module is trained using textual and visual prompts retrieved from caches to learn a multimodal category prompt.

\begin{table*}[t]
\setlength{\tabcolsep}{8pt}
  \centering
  \caption{
    Comparison with existing open-vocabulary detection methods.
Results for GLIP, GroundingDINO, and LAE-DINO are taken from the LAE-DINO paper, while OpenRSD results are from its original paper.
DIOR is evaluated using AP$_{50}$, and DOTA-v2.0 and LAE-80C are evaluated using mAP (AP$_{50}$ is additionally reported for DOTA-v2.0).
Unless otherwise specified, visual prompting uses 32 visual prompts, and multimodal prompting combines textual prompts with 32 visual prompts.
  }
  \label{tab:sota-config}
  \begin{tabular}{l l l c c c}
    \toprule
    Method & Pretrain Dataset & Prompt &
    \makecell{DIOR \\ (AP$_{50}$)} &
    \makecell{DOTA-v2.0 \\ (mAP / AP$_{50}$)} &
    \makecell{LAE-80C \\ (mAP)} \\
    \midrule
    GLIP \cite{li2022glip}                       & LAE-1M  & text            & 82.8 & 43.0 / --   & 16.5 \\
    GroundingDINO \cite{liu2024gdino}              & LAE-1M  & text            & 83.6 & 46.0 / --   & 17.7 \\
    LAE-DINO \cite{pan2025laedino}                   & LAE-1M  & text            & \textbf{85.5} & 46.8 / -- & 20.2 \\
    OpenRSD \cite{huang2025openrsd}                   & ORSD+   & text            & 76.7 & -- / 71.8   & --   \\
    OpenRSD \cite{huang2025openrsd}                   & ORSD+   & visual          & 76.7 & -- / 69.8   & --   \\
    \midrule
    RS-MPOD (Ours)              & LAE-1M  & text            & 81.3 & 46.1 / 73.8 & 23.1 \\
    RS-MPOD (Ours)            & LAE-1M  & visual–32       & 76.2 & 44.1 / 70.1 & 20.1 \\
    RS-MPOD (Ours)   & LAE-1M  & multimodal  &
                                 83.6 &
                                 \textbf{48.0} / \textbf{74.6} &
                                 \textbf{24.1} \\
    \bottomrule
  \end{tabular}
\end{table*}

\begin{table*}[t]
  \centering
  \caption{
    Zero-shot cross-dataset generalization results on remote sensing benchmarks. 
    Models are evaluated using AP$_{50}$ under different prompting configurations.
  }
  \label{tab:cross-dataset-config}
  \begin{tabular}{l l c c c c c}
    \toprule
    Method & Prompt &
    HRRSD & VisDrone & AI-TOD & LEVIR \\
    \midrule
    GroundingDINO \cite{liu2024gdino} & text & 43.7 & 4.7 & 30.1  & 49.6 \\
    LAE-DINO \cite{pan2025laedino} & text & 42.8 & 2.3 & 31.5 & 50.8 \\
    RS-MPOD (Ours) & text               & 41.9 & 3.8 & 27.6  & 48.2 \\
    RS-MPOD (Ours) & visual–32          & 43.2 & \textbf{5.5} & \textbf{33.7}  & 46.9 \\
    RS-MPOD (Ours) & multimodal     &
            \textbf{44.1} & 4.6 & 32.8 &  \textbf{51.0} \\
    \bottomrule
  \end{tabular}
\end{table*}

\begin{table*}[t]
  \centering
  \caption{
    Zero-shot cross-dataset generalization results on fine-grained remote sensing benchmarks.
    All models are evaluated using AP$_{50}$ under different prompting configurations.
  }
  \label{tab:finegrained-cross-dataset}
  \begin{tabular}{l l c c c c c}
    \toprule
    Method & Prompt &
    SIMD & MVRSD & MAR20 & FGSD2021 \\
    \midrule
    GroundingDINO \cite{liu2024gdino} & text &  17.7 & 10.4 & 5.7 & 7.7\\
    LAE-DINO \cite{pan2025laedino} & text &  20.5 & 4.6 & 3.7 & 4.7\\
    RS-MPOD (Ours) & text               &   16.4 & 1.0 & 4.6 & 8.0\\
    RS-MPOD (Ours) & visual–32          &  \textbf{30.6} & \textbf{22.0} & \textbf{11.2} & \textbf{14.7}\\
    RS-MPOD (Ours) & multimodal     &
              29.2 & 19.5 & 6.7 & 7.0\\
    \bottomrule
  \end{tabular}
\end{table*}

\section{Experiments}
\subsection{Datasets}

We adopt LAE-1M~\cite{pan2025laedino} as the primary dataset for training.
LAE-1M contains over 1,600 object categories and is a large-scale remote sensing dataset designed for open-vocabulary object detection.
It consists of two components:
(i) \textbf{LAE-FOD}, which is curated from existing remote sensing detection datasets and annotated with high-quality human labels; and
(ii) \textbf{LAE-COD}, which is automatically constructed from unlabeled image--text pairs through an automated pipeline, expanding the category set beyond manually annotated data.

\subsection{Evaluation Benchmarks}

We evaluate our method on multiple remote sensing detection benchmarks covering in-domain evaluation, cross-dataset generalization, and fine-grained category recognition.
Specifically, DIOR~\cite{li2020dior} and DOTA-v2.0~\cite{xia2018dota} are used as representative benchmarks for in-domain open-vocabulary detection, while LAE-80C~\cite{pan2025laedino} is adopted to assess performance under a substantially larger category space.
Cross-dataset generalization is evaluated on HRRSD~\cite{zhang2019hrrsd}, VisDrone~\cite{zhu2021visdrone}, AI-TOD~\cite{wang2021aitod}, and LEVIR~\cite{zou2017levir}, which cover diverse sensing conditions and scene distributions.
In addition, we evaluate open-vocabulary object detection on fine-grained remote sensing benchmarks, including SIMD~\cite{haroon2020simd}, MVRSD~\cite{bai2024mvrsd}, MAR20~\cite{wenqi2024mar20}, and FGSD2021~\cite{zhang2021fgsd}, where categories exhibit subtle visual differences and more specialized category definitions.
All evaluations are conducted using horizontal bounding boxes.
AP$_{50}$ is reported on DIOR and all cross-dataset and fine-grained benchmarks, while mAP is used on LAE-80C; for DOTA-v2.0, both mAP and AP$_{50}$ are reported for fair comparison with prior work.

\subsection{Implementation Details}

We follow the three-stage training pipeline described in Sec.~\ref{sec:stagewise-training}.
Stages~I and II are trained on the full LAE-1M dataset, including both LAE-FOD and LAE-COD.
In Stage~I, the text-prompt-based detector is trained for 24 epochs.
In Stage~II, the detector is frozen and only the visual prompt encoder is trained for 12 epochs.
Stage~III trains only the multimodal prompt fusion module for 12 epochs and uses the LAE-FOD subset, as visual prompts constructed from sparse or noisy categories in LAE-COD are less suitable for fusion training.
All experiments are conducted using four NVIDIA RTX 3090 GPUs with distributed training.

\subsection{Main Results}

\noindent\textbf{Comparison with existing methods.}
Table~\ref{tab:sota-config} compares our approach with representative open-vocabulary detectors.
We include general-domain methods such as GLIP~\cite{li2022glip} and GroundingDINO~\cite{liu2024gdino}, as well as remote-sensing--specific approaches including LAE-DINO~\cite{pan2025laedino} and OpenRSD~\cite{huang2025openrsd}.
This evaluation focuses on standard in-domain benchmarks, where category definitions and annotation protocols are consistent with the training data.
Under text-only prompting, our model achieves 81.3\% AP$_{50}$ on DIOR, 46.1\% mAP on DOTA-v2.0, and 23.1\% mAP on LAE-80C.
Compared with LAE-DINO, this corresponds to slightly lower performance on DIOR and DOTA-v2.0, while yielding higher mAP on LAE-80C.
With 32 visual prompts, our model reaches 76.2\% AP$_{50}$ on DIOR and 70.1\% AP$_{50}$ on DOTA-v2.0, achieving performance comparable to the visual-prompt results reported by OpenRSD.
The multimodal configuration, which combines textual prompts with 32 visual prompts, provides the most stable overall performance across benchmarks.
It obtains 48.0\% mAP on DOTA-v2.0 and 24.1\% mAP on LAE-80C, exceeding LAE-DINO by +1.2\% and +3.9\% mAP, respectively.
On DIOR, multimodal prompting reaches 83.6\% AP$_{50}$, matching the performance of GroundingDINO and remaining close to LAE-DINO.
Overall, these results indicate that incorporating visual and multimodal prompting maintains competitive performance on standard benchmarks, while providing consistent gains when complementary category cues are available.

\begin{table}[t]
\setlength{\tabcolsep}{9pt}
  \centering
  \caption{
    Ablation study on different prompting configurations.
    DIOR is evaluated using AP$_{50}$, while DOTA-v2.0 and LAE-80C are evaluated using mAP.
    ``visual--N'' denotes the use of N visual prompts aggregated into a single category representation at inference time.
  }
  \label{tab:prompt-config}
  \begin{tabular}{l c c c}
    \toprule
    Prompt &
    \makecell{DIOR \\ (AP$_{50}$)} & \makecell{DOTA-v2.0 \\ (mAP)} & \makecell{LAE-80C \\(mAP)} \\
    \midrule
     text                 & 81.3 & 46.1 & 23.1 \\

      visual–1                & 65.5 & 37.3 & 14.3 \\
      visual–4                & 73.6 & 43.4 & 17.9 \\
      visual–8                & 76.4 & 43.7 & 19.3 \\
      visual–16               & 75.7 & 44.2 & 19.6 \\
      visual–32               & 76.2 & 44.1 & 20.1 \\

      multimodal       &
                \textbf{83.6} & \textbf{48.0} & \textbf{24.1} \\
    \bottomrule
  \end{tabular}
\end{table}

\begin{table}[t]
\setlength{\tabcolsep}{3pt} 
\centering
\caption{
Effect of freezing the detector during visual prompt encoder training.
``Unfrozen'' and ``Frozen'' indicate whether detector parameters are updated during training. 
}
\label{tab:ablation-freeze}
\begin{tabular}{l c c c}
\toprule
Condition & Prompt &
DIOR (AP$_{50}$) &
DOTA-v2.0 (mAP) \\
\midrule
Unfrozen  & visual–32 & 60.9 & 34.9  \\
Frozen     & visual–32 & \textbf{76.2} & \textbf{44.1} \\
\bottomrule
\end{tabular}
\end{table}

\begin{table}[t]
\setlength{\tabcolsep}{1pt}
\centering
\caption{
Comparison of fusion strategies for combining textual and visual prompts.
The ``Avg'' baseline represents simple averaging of textual and visual prompts, 
whereas ``Fusion'' denotes our proposed module for learned multimodal integration.
}
\label{tab:ablation_fusion_avg}
\begin{tabular}{l c c c}
\toprule
Strategy & Prompt &
DIOR (AP$_{50}$) &
DOTA-v2.0 (mAP) \\
\midrule
Avg    & multimodal & 79.8 & 45.4 \\
Fusion & multimodal & \textbf{83.6} & \textbf{48.0} \\
\bottomrule
\end{tabular}
\end{table}

\noindent\textbf{Cross-dataset generalization.}
Table~\ref{tab:cross-dataset-config} reports zero-shot detection results on several external datasets.
Several consistent patterns can be observed across cross-dataset settings.
Within RS-MPOD, visual prompting achieves higher or more stable performance than textual prompting on most benchmarks, with particularly notable gains on AI-TOD (+6.1\% AP$_{50}$).
On these datasets, RS-MPOD with visual prompting also achieves performance comparable to or exceeding representative text-prompt-based methods such as GroundingDINO and LAE-DINO.
This behavior differs from the in-domain results, where textual prompting generally performs better.
A consistent observation is that the effectiveness of textual prompting varies substantially across datasets, suggesting that inference-time category specifications may not consistently align with the text--visual alignment learned during pretraining under distribution shifts.
By contrast, visual prompts rely on appearance cues extracted from exemplar instances to specify categories based on visual similarity.
Multimodal prompting provides additional gains on some datasets, such as HRRSD and LEVIR (e.g., +2.8\% AP$_{50}$ on LEVIR).
On VisDrone and AI-TOD, multimodal prompting performs comparably to or slightly below visual-only prompting, indicating that instance-level appearance cues play a dominant role in cross-dataset generalization in these cases, particularly when category semantics differ substantially across datasets.

\noindent\textbf{Fine-grained cross-dataset generalization.}
Table~\ref{tab:finegrained-cross-dataset} reports zero-shot detection results on several fine-grained remote sensing benchmarks.
Compared with the general cross-dataset setting, fine-grained benchmarks exhibit a substantially larger performance gap between textual and visual prompting.
Across all fine-grained datasets, visual prompting achieves consistently higher performance than text-only prompting, with particularly large gains of +14.2\% AP$_{50}$ on SIMD, +21.0\% AP$_{50}$ on MVRSD, +6.6\% AP$_{50}$ on MAR20, and +6.7\% AP$_{50}$ on FGSD2021.
On these benchmarks, RS-MPOD with visual prompting also substantially outperforms representative text-prompt-based methods such as GroundingDINO and LAE-DINO.
Multimodal prompting remains competitive across all benchmarks, while performing comparably to or slightly below visual-only prompting on most fine-grained datasets, as poorly aligned textual cues can introduce additional semantic noise during fusion.
Overall, these results indicate that under fine-grained cross-dataset evaluation, discrepancies between pretraining semantics and inference-time category definitions become more pronounced, under which category specification based on instance-level appearance cues remains more stable than text-only prompting.

\subsection{Ablation Experiments}
\noindent\textbf{Results under different prompting configurations.}
Table~\ref{tab:prompt-config} reports the performance under different prompting configurations.
Several consistent patterns can be observed across benchmarks.
First, using a single visual prompt results in notably lower performance, indicating that a single exemplar provides limited coverage of intra-class appearance variations.
Second, performance improves steadily as the number of visual prompts increases: increasing the number of visual prompts from Visual-1 to Visual-32 leads to gains of +10.7\% AP$_{50}$ on DIOR, +6.8\% mAP on DOTA-v2.0, and +5.8\% mAP on LAE-80C.
This trend suggests that aggregating multiple exemplar instances enables the visual prompt encoder to form more representative category-level appearance cues.
Third, even with multiple visual prompts, the visual-only configuration performs slightly below the text-only baseline.
This difference reflects the distinct roles of textual and visual prompting in category specification.
Textual prompts provide category identifiers that directly correspond to benchmark labels, whereas visual prompts rely on a limited set of exemplar appearances, which may be insufficient to fully characterize category boundaries.
Finally, the multimodal configuration consistently achieves the strongest performance across benchmarks, outperforming the text-only baseline by +2.3\% AP$_{50}$ on DIOR, +1.9\% mAP on DOTA-v2.0, and +1.0\% mAP on LAE-80C.
These results confirm that textual and visual prompts provide complementary category cues, and their integration leads to more effective category specification.

\noindent\textbf{Effect of freezing the detector during visual prompt encoder training.}
Table~\ref{tab:ablation-freeze} examines the effect of freezing the detector when training the visual prompt encoder.
In the first configuration, the detector and the visual prompt encoder are jointly optimized, with each iteration conditioned on either textual or visual prompts.
In the second configuration, all detector parameters are frozen and only the visual prompt encoder is updated.
Freezing the detector leads to substantially better performance, improving DIOR AP$_{50}$ by +15.3\% and DOTA-v2.0 mAP by +9.2\%.
This result indicates that learning visual prompts benefits from a stable category-querying space.
When the detector continues to update, both visual and query representations change simultaneously, making it difficult for the visual prompt encoder to adapt stably.
By freezing the detector, the visual prompt encoder is trained with fixed visual representations and a stable querying mechanism, enabling more effective instance-grounded prompt embeddings.

\noindent\textbf{Effect of fusion strategy for combining textual and visual prompts.}
Table~\ref{tab:ablation_fusion_avg} compares the proposed multimodal prompt fusion module with a non-learnable baseline.
In the baseline, the visual prompt for each category is directly added to the token-level textual features, and the resulting features are averaged to obtain a category representation.
Under the same textual+Visual-32 inference setting, the proposed fusion module consistently outperforms the non-learnable baseline, improving DIOR AP$_{50}$ by +3.8\% and DOTA-v2.0 mAP by +2.6\%.
This result suggests that effectively combining textual and visual prompts requires a dedicated fusion mechanism that accounts for their different structures and roles in category specification, rather than treating them as interchangeable features through averaging.

\section{Conclusion}
In this work, we focus on category specification in open-vocabulary object detection for remote sensing, where inference-time category definitions and granularity often differ from training data, challenging the reliability of text-only prompting.
Our experiments show that although text-only prompting can perform well when inference-time category semantics align with learned text--visual correspondence, its effectiveness degrades under cross-dataset shifts and becomes particularly fragile in fine-grained settings.
To address this limitation, we propose RS-MPOD, a multimodal prompting framework that enables category specification through instance-grounded visual exemplars and their integration with textual prompts.
Results across in-domain, cross-dataset, and fine-grained benchmarks demonstrate that visual prompting yields more stable category cues under semantic or distribution shifts, while multimodal prompting provides a flexible and robust alternative when both textual and visual cues are available.

\section*{Impact Statement}

This paper presents work whose goal is to advance the field of Machine Learning. There are many potential societal consequences of our work, none which we feel must be specifically highlighted here.

\nocite{langley00}

\bibliography{main}
\bibliographystyle{icml2026}

\newpage
\appendix
\onecolumn
\section{Appendix for Experiments.}

\subsection{Fine-Grained Category Definitions and Mappings}
This appendix summarizes the category definitions of the fine-grained remote sensing datasets used in our experiments, including SIMD~\cite{haroon2020simd}, MVRSD~\cite{bai2024mvrsd}, MAR20~\cite{wenqi2024mar20}, and FGSD2021~\cite{zhang2021fgsd}.
These datasets adopt specialized or application-specific taxonomies that are not commonly used in standard object detection benchmarks.
For clarity and reproducibility, we list the category names of all fine-grained datasets in Table~\ref{tab:finegrained-categories}.
In addition, for datasets with non-semantic identifiers or overly specialized class names that are difficult to specify using textual prompts, we adopt category name normalization or functional grouping for analysis.
These modifications are limited to category naming for semantic interpretability and textual prompting, while the original annotations, instance assignments, and evaluation protocols remain unchanged.

\begin{table}[h]
\centering
\caption{Category definitions of fine-grained remote sensing datasets used in this work.}
\label{tab:finegrained-categories}
\begin{tabular}{ll}
\toprule
Dataset & Category Names \\
\midrule
\makecell[l]{SIMD } &
\makecell[l]{Van, Long Vehicle, Bus, Airliner, Propeller Aircraft, Trainer Aircraft, Chartered Aircraft, \\Fighter Aircraft, Helicopter, Boat, Stair Truck, Pushback Truck, Car, Truck, Others} \\
\midrule
\makecell[l]{MVRSD } &
\makecell[l]{Small Military Vehicles, Large Military Vehicles, Armored Fighting Vehicles, \\Military Construction Vehicles, Civilian Vehicles} \\
\midrule
\makecell[l]{MAR20 } &
\makecell[l]{Su-35 Fighter, Tu-160 Bomber, Tu-22 Bomber, Tu-95 Bomber, Su-34 Fighter-Bomber,\\ Su-24 Fighter-Bomber, C-130 Transport Aircraft, C-17 Transport Aircraft, F-22 Fighter,\\ F-16 Fighter, E-3 AWACS, B-52 Bomber, P-3C Anti-Submarine Aircraft, B-1B Bomber,\\ E-8 JSTARS, F-15 Fighter, KC-135 Aerial Refueling Aircraft, C-5 Transport Aircraft,\\ F/A-18 Fighter-Attack Aircraft, KC-10 Aerial Refueling Aircraft} \\
\midrule
\makecell[l]{FGSD2021} &
\makecell[l]{Aircraft Carrier, Cruiser, Destroyer, Frigate, Amphibious Ship, Support Ship, Submarine, Other} \\
\bottomrule
\end{tabular}
\end{table}

\begin{table}[h]
\centering
\caption{Mapping from original class names to updated category names for fine-grained aircraft and ship datasets.}
\label{tab:aircraft-ship-sidebyside}
\begin{tabular}{ll|ll}
\toprule
\multicolumn{2}{c|}{MAR20 (Aircraft)} & \multicolumn{2}{c}{FGSD2021 (Ships)} \\
\midrule
Original Class Name & Updated Category Name & Original Class Name & Updated Category Name \\
\midrule
A1  & Su-35 Fighter & Ticonderoga-class & Cruiser \\
A2  & Tu-160 Bomber & Perry-class & Frigate \\
A3  & Tu-22 Bomber & Freedom-class & Frigate \\
A4  & Tu-95 Bomber & Independence-class & Frigate \\
A5  & Su-34 Fighter-Bomber & Arleigh Burke-class & Destroyer \\
A6  & Su-24 Fighter-Bomber & Aircraft Carrier & Aircraft Carrier \\
A7  & C-130 Transport Aircraft & Submarine & Submarine \\
A8  & C-17 Transport Aircraft & Tarawa-class & Amphibious Ship \\
A9  & C-5 Transport Aircraft & Austin-class & Amphibious Ship \\
A10 & F-16 Fighter & Wasp-class & Amphibious Ship \\
A11 & E-3 AWACS & Whidbey Island-class & Amphibious Ship \\
A12 & B-52 Bomber & San Antonio-class & Amphibious Ship \\
A13 & P-3C Anti-Submarine Aircraft & Newport-class & Amphibious Ship \\
A14 & B-1B Bomber & Kaiser-class & Support Ship \\
A15 & E-8 JSTARS & Avenger-class & Support Ship \\
A16 & F-15 Fighter & Hope-class & Support Ship \\
A17 & KC-135 Aerial Refueling Aircraft & Supply-class & Support Ship \\
A18 & F-22 Fighter & Mercy-class & Support Ship \\
A19 & F/A-18 Fighter-Attack Aircraft & Lewis and Clark-class & Support Ship \\
A20 & KC-10 Aerial Refueling Aircraft & Other & Other \\
\bottomrule
\end{tabular}
\end{table}

Both MAR20 and FGSD2021 contain highly fine-grained categories with specialized naming conventions.
For MAR20, abstract identifiers are replaced with semantically meaningful aircraft type names, while FGSD2021 ship classes are grouped into broader functional categories for analysis, as summarized in Table~\ref{tab:aircraft-ship-sidebyside}.
These adjustments affect only category name representation and do not alter the underlying annotation structure or evaluation procedures.

\subsection{Appendix for Ablation Experiments.}

\noindent\textbf{Effect of the number of sampled instances for constructing visual prompts.}
Table~\ref{tab:ablation-vp-boxes} analyzes how the number of sampled instances per category during training influences the performance of the visual prompt encoder.
Across different inference configurations, sampling a single instance per category in each training iteration consistently yields the strongest performance on both DIOR and DOTA-v2.0.
Increasing the number of sampled instances to four or eight per category does not provide additional gains and in some cases leads to slightly lower AP$_{50}$/mAP.
This behavior suggests that sampling a single instance introduces higher variability in prompt construction across training iterations, whereas sampling multiple instances simultaneously results in more stable but less diverse prompts.
Such increased variability may serve as an implicit regularization mechanism, improving the robustness of the learned instance-grounded visual prompts.

\begin{table}[h]
\centering
\caption{
Effect of the number of sampled instances per category during training for visual prompt construction.
}
\label{tab:ablation-vp-boxes}
\begin{tabular}{lccc|ccc}
\toprule
\multirow{2}{*}{Prompt} &
\multicolumn{3}{c}{DIOR (AP$_{50}$)} &
\multicolumn{3}{c}{DOTA-v2.0 (mAP)} \\
\cline{2-4} \cline{5-7}
& 1 & 4 & 8 & 1 & 4 & 8 \\
\midrule
visual–1   & \textbf{65.5} & 59.3 & 57.4 & \textbf{37.3} & 35.8 & 32.8 \\
visual–4   & \textbf{73.6} & 70.6 & 72.7 & \textbf{43.4} & 40.2 & 42.4 \\
visual–8   & \textbf{76.4} & 75.8 & 74.5 & \textbf{43.7} & 41.9 & 42.2 \\
visual–16  & \textbf{75.7} & 75.3 & 75.6 & \textbf{44.2} & 43.5 & 43.9 \\
visual–32  & \textbf{76.2} & 75.5 & 76.1 & \textbf{44.1} & 43.5 & 43.4 \\
\bottomrule
\end{tabular}
\end{table}

\noindent\textbf{Effect of the number of visual prompts used for training the fusion module.}
Table~\ref{tab:ablation_vp_num} examines how the number of visual prompts provided during training affects the multimodal prompt fusion module.
We consider three training strategies: using a single visual prompt, randomly sampling a variable number of prompts, and using the full set of 32 visual prompts.
All models are evaluated under the same textual+Visual-32 inference configuration.
Training with the full set of 32 visual prompts consistently yields the best performance across benchmarks.
Reducing the number of visual prompts during training results in lower AP$_{50}$/mAP, indicating that exposing the fusion module to a broader and more diverse set of visual prompts is important for learning effective multimodal category representations.

\begin{table}[h]
\setlength{\tabcolsep}{3pt}
\centering
\captionsetup{width=0.7\linewidth,justification=centering}
\caption{
Effect of the number of visual prompts used to train the multimodal fusion module.
During training, ``visual--r'' denotes random sampling of visual prompts.
}
\label{tab:ablation_vp_num}
\begin{tabular}{l l c c}
\toprule
\begin{tabular}{@{}c@{}}Training \\ prompt\end{tabular} &
\begin{tabular}{@{}c@{}}Inference \\ prompt\end{tabular} &
DIOR (AP$_{50}$) & DOTA-v2.0 (mAP) \\
\midrule
visual–1   & visual–32 & 81.4 & 47.3 \\
visual–r   & visual–32 & 83.0 & 47.6 \\
visual–32  & visual–32 & \textbf{83.6} & \textbf{48.0} \\
\bottomrule
\end{tabular}
\end{table}

\section{Appendix for Results Visualization.}

\begin{figure*}
    \centering
    \includegraphics[width=0.95\linewidth]{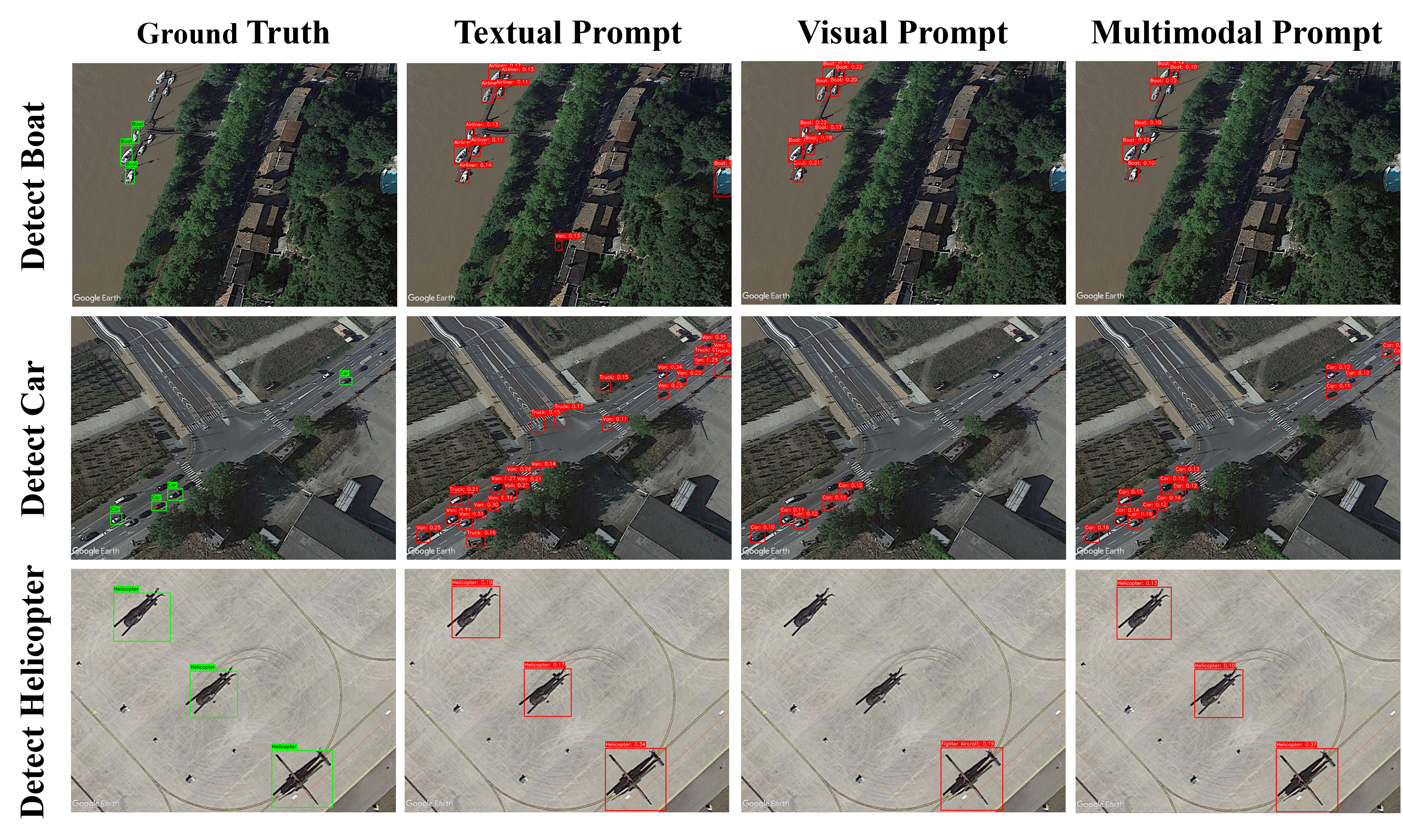}
    \caption{
    Qualitative comparison of detection results under different prompting strategies.
    From left to right: ground truth, text-only prompting, visual-only prompting, and multimodal prompting.
    The three rows illustrate representative scenarios with varying degrees of semantic alignment between pretraining and evaluation: (top) ship detection with category name mismatch (boat vs.\ ship), (middle) car detection with weak textual representation and confusion with visually similar categories (e.g., van), and (bottom) helicopter detection with strong semantic alignment in pretraining.
    Multimodal prompting exhibits more balanced behavior across scenarios by combining complementary textual and visual cues.
    }
    \label{fig:qualitative_prompting}
\end{figure*}

Figure~\ref{fig:qualitative_prompting} provides a qualitative comparison of detection results under different prompting strategies, illustrating how textual, visual, and multimodal prompts behave under varying semantic conditions.
In the first row, ship detection results are shown for a scenario where the evaluation dataset uses the category name \emph{boat}, while large-scale pretraining corpora predominantly employ the term \emph{ship}.
Under text-only prompting, this semantic mismatch leads the detector to confuse boats with visually related categories such as \emph{airliner}, whereas visual prompting correctly localizes and classifies the targets by relying on instance-level appearance cues.
In the second row, the model is evaluated on car detection in a setting where the concept of \emph{car} is weakly represented in pretraining data, but visually similar categories such as \emph{van} are more prevalent.
As a result, text-only prompting tends to misclassify cars as vans, while visual prompting reduces this confusion by grounding category specification in visual exemplars.
In contrast, the third row shows helicopter detection in a scenario where textual category semantics are well aligned with pretraining, leading to strong performance under text-only prompting, while visual prompting alone fails to detect some instances due to limited exemplar coverage.
Across all scenarios, multimodal prompting achieves more balanced performance by integrating complementary textual and visual cues, demonstrating improved adaptability across diverse semantic conditions.

\end{document}